\documentclass{article}

\usepackage{microtype}
\usepackage{graphicx}

\usepackage[export]{adjustbox}

\usepackage{hyperref}

\usepackage[accepted]{icml2024mi}

\usepackage{amsmath}
\usepackage{amssymb}
\usepackage{mathtools}

\usepackage[capitalize,noabbrev]{cleveref}

\icmltitlerunning{Penzai + Treescope: A Toolkit for Interpreting, Visualizing, and Editing Models As Data}

\begin{document}

\twocolumn[
\icmltitle{Penzai + Treescope:\\A Toolkit for Interpreting, Visualizing, and Editing Models As Data}

\icmlsetsymbol{equal}{*}

\begin{icmlauthorlist}
\icmlauthor{Daniel D. Johnson}{gdm,uoft}
\end{icmlauthorlist}

\icmlaffiliation{gdm}{Google DeepMind}
\icmlaffiliation{uoft}{University of Toronto, Department of Computer Science, Ontario, Canada}

\icmlcorrespondingauthor{Daniel D. Johnson}{ddjohnson@cs.toronto.edu}

\icmlkeywords{Machine Learning, ICML}

\vskip 0.3in
]

\printAffiliationsAndNotice{}  %

\begin{abstract}%
Much of today's machine learning research involves interpreting, modifying or visualizing models after they are trained.
I present \emph{Penzai}, a neural network library designed to simplify model manipulation by representing models as simple data structures, and \emph{Treescope}, an interactive pretty-printer and array visualizer that can visualize both model inputs/outputs and the models themselves.
Penzai models are built using declarative combinators that expose the model forward pass in the structure of the model object itself, and use named axes to ensure each operation is semantically meaningful.
With Penzai's tree-editing selector system, users can both insert and replace model components, allowing them to intervene on intermediate values or make other edits to the model structure. Users can then get immediate feedback by visualizing the modified model with Treescope.
I describe the motivation and main features of Penzai and Treescope, and discuss how treating the model as data enables a variety of analyses and interventions to be implemented as data-structure transformations, without requiring model designers to add explicit hooks.
\end{abstract}

\section{Introduction}
Due to the increasing capabilities of large language models and other foundation models, and the similarly increasing cost to training them, much research with large models has shifted to after their initial training run. This includes interpreting the ``circuits'' inside models \citep[e.g.][]{wang2022interpretability}, probing internal representations \citep[e.g.][]{luo2021local}, or fine-tuning models using parameter-efficient adapters to control their behavior \citep[e.g.][]{hu2021lora}. Conducting this research often involves either \emph{visualizing} model components, \emph{inserting} new logic to intervene on activations, \emph{replacing} individual model components, or some combination of these.

Unfortunately, the original model representation is often not well-suited for making modifications, as most model codebases are designed for efficient training and inference. And working around these limitations often comes at the expense of readability or missing functionality. As one example, TransformerLens \citep{nanda2022transformerlens} supports a wide set of interventions using a ``hooked'' reimplementation of common transformer variants, but does not support efficient sampling or multiple-device computation. 
This kind of analysis has historically been even more complex when using JAX \citep{jax2018github}, because hook-based interfaces use global state that is  difficult to combine with JAX's purely functional transformations.

Similarly, the original model representation is not usually designed to enable easy visualization.
Existing tools, such as the Language Interpretability Tool \citep{tenney2020language} or the Transformer Debugger neuron viewer \citep{mossing2024tdb}, have been introduced to help researchers understand model behavior, but tend to support visualizations of only model outputs or specific intermediate values (such as attention patterns) and do not support visualizing the model structure itself. General-purpose plotting libraries like Matplotlib \citep{Hunter:2007} or Plotly \citep{plotly} tend to prioritize tabular data and are not well-suited to visualizing data involving multidimensional arrays.

\begin{figure*}[t]
    \centering
    \includegraphics[width=0.99\linewidth,trim={0 0 2in 0},clip]{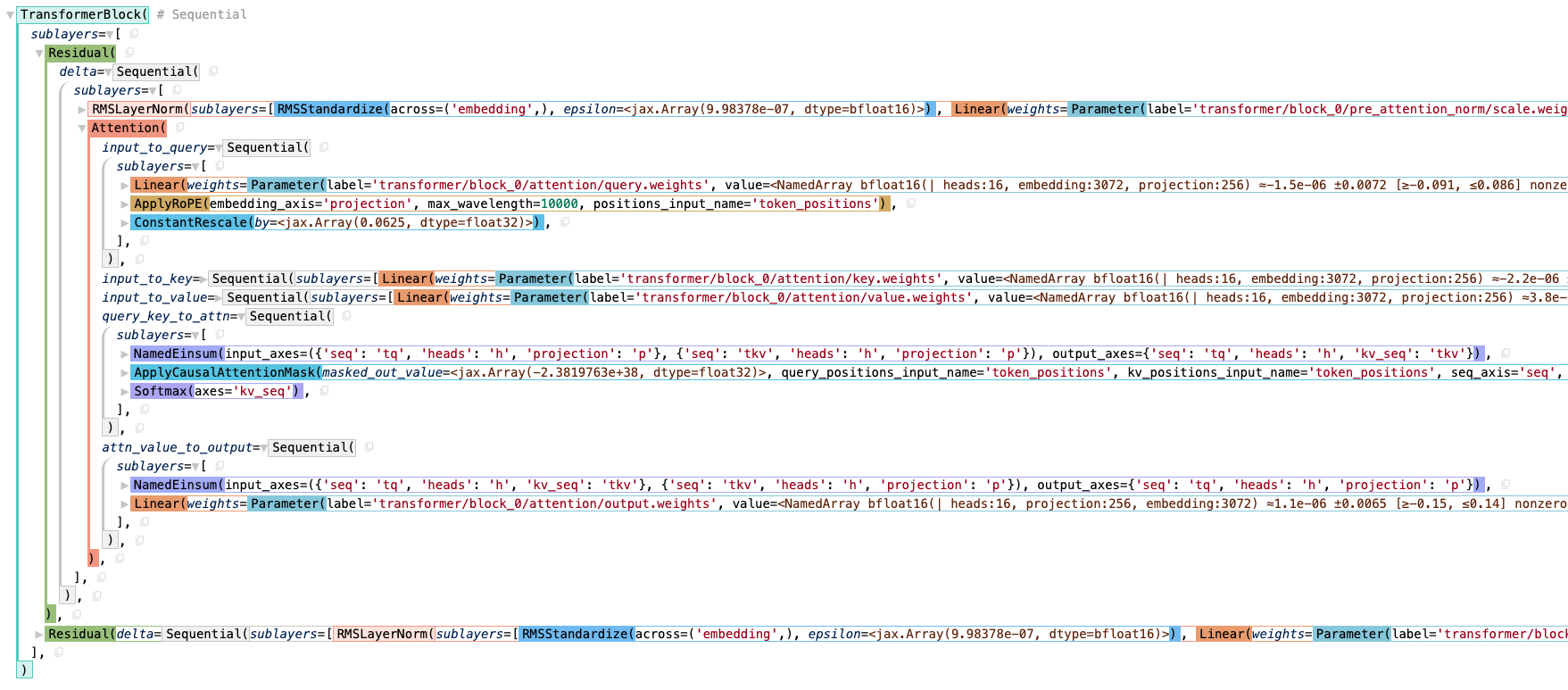}
    \vspace{-0.5em}
    \caption{A partially-expanded Treescope rendering of a Transformer block from Penzai's implementation of the Gemma 7B model \citep{team2024gemma}, showing the \texttt{pz.nn.Attention} combinator and some of the primitive sublayers it contains.}
    \label{fig:treescope-combinators}
\end{figure*}

This paper describes Penzai, a JAX library focused on making it easier to manipulate complex neural network models and their activations, and Treescope, a Python pretty-printer designed to visualize models and multidimensional-array data. These libraries aim to simplify research into pretrained models by first \emph{treating models as simple data structures}, which are designed to be modular and can be directly manipulated by the user in order to change their behavior, and then \emph{providing general tools for visualizing and editing those data structures} in an interactive setting. This allows Penzai and Treescope to support a ``what-you-see-is-what-you-get'' research workflow: model interventions can be visualized simply by pretty-printing the modified model, and there is always a direct correspondence between the model's internal structure, the structure of its pretty-printed visualization, and its runtime behavior.
Penzai and Treescope are open-source, and available at \texttt{\small\href{https://github.com/google-deepmind/penzai}{github.com/google-deepmind/penzai}} and \texttt{\small\href{https://github.com/google-deepmind/treescope}{github.com/google-deepmind/treescope}}.

\section{Previous Model-Manipulation Strategies}
A number of libraries have proposed interfaces for intervening on model activations while a model runs.
TransformerLens \citep{nanda2022transformerlens} includes a transformer implementation with \textbf{hook points}, and enables users to add hook functions that can read or modify activations. This supports a wide variety of transformations, but also requires users to manage global state of hooks, and does not support efficient sampling or multi-device computation.
\texttt{pyvene} \citep{wu2024pyvene} similarly allows modifying internal model activations with new logic, but represents interventions using a \textbf{intervention schema} instead of global hooks. NNSight \citep{nnsight} uses a \textbf{tracing context} to provide an interface where activations of PyTorch models can be extracted and modified based on their location in the model; these modifications are then converted to a graph and evaluated on remote workers.
A common difficulty with many of these approaches is that activations must first be located before they can be modified; each transformer implementation and modification library comes with its own conventions and syntax for accessing them.

A different form of model modification is \textbf{hot-swapping}, where parts of a model are replaced entirely with other parts. This strategy is sometimes used to add new parameters to pretrained models for parameter-efficient fine-tuning. For instance, the PEFT \citep{peft} and Levanter \citep{levanter} libraries replace internal linear layers with low-rank adapter layers to implement LoRA fine-tuning \citep{hu2021lora}, although the details of this are mostly abstracted away from the user.

\section{Penzai: Treating the Forward Pass as Data}

How can we make it as easy as possible for new users to inspect and modify the behavior of models that they did not train? The central idea in Penzai is to focus on decomposing the model object into self-explanatory pieces, and allow the user to directly modify and recombine these pieces in new ways. This removes the need for hooks or tracing, and makes it easy to tell how to change the behavior of a model: simply modify the model object itself.

\subsection{Combinators and Primitive Layers}
Concretely, Penzai provides a library of simple neural network components that can be combined to implement more complex models. Many of these components are \emph{combinators}, which use child layers to implement more complex behaviors. This includes standard combinators such as \texttt{Sequential} (also provided by libraries like PyTorch and Keras), but also more advanced combinators such as \texttt{BranchAndMultiplyTogether}, which multiplies the results of running its children in parallel, and \texttt{Attention}, which routes intermediates between query, key, and value heads. Importantly, these combinators make minimal assumptions about their children. For instance, the \texttt{Attention} combinator has no parameters and is just responsible for routing queries, keys, values, and outputs, not for computing dot products or attention masks; those are implemented by child layers. This means that the same \texttt{Attention} combinator is compatible with many common attention variants, such as rotary positional embeddings \citep{su2024roformer} or grouped-query attention \citep{ainslie2023gqa}, without requiring a complex implementation.
In addition to these combinators, Penzai also provides components for primitive operations such as \texttt{Linear}, \texttt{AddBias}, \texttt{ApplyCausalAttentionMask}, and \texttt{Softmax}. Primitive operations are written in terms of named axes instead of axis positions, to ensure their behavior does not require memorizing axis ordering conventions.

To keep model structures as simple as possible, Penzai avoids storing configuration data (such as activation functions or causal decoding flags) as attributes on the model object, and models do not use conditional branching at runtime. Instead, each model's sublayers are specialized for a single configuration, and store only the information that they need, using fully-documented and type-annotated attributes. By convention, each layer accepts a single ``main input'', usually the activations from the previous layer, along with a set of ``side input'' keyword arguments that are shared across all layers and provide context such as token positions or attention masks. This convention makes it possible to compose layers together in a generic way.

Users can customize the behavior of models or change their configuration by directly hot-swapping different implementations, e.g. replacing each \texttt{Attention} combinator with a \texttt{KVCachingAttention} combinator to enable fast autoregressive decoding.
This means model implementations in Penzai have a one-to-one correspondence between the Python structure of the model object and the computations that run during its forward pass. Users are free to change how their model works without using hooks or intervention schemas, by instead directly inserting a new primitive into the model at the relevant place, and using pretty-printing to identify which locations they need to modify. An example Transformer block in Penzai is shown in \cref{fig:treescope-combinators}.

\subsection{Lightweight Named Axes System}
Giving axes names makes it easier to identify the purpose of each axis, prevents users from needing to memorize specific ordering conventions, and simplifies visualizing array data. Unfortunately, named axis implementations often impose an additional implementation burden due to changing the semantics of each individual operation.
To give the benefits of named axes without requiring changes to the array API, Penzai includes a lightweight named axis system based on \emph{locally-positional semantics} and inspired by \citet{chiang2021named}. Users interact with this system by constructing positional \emph{views} of a subset of axes in an array and then applying ordinary JAX operations to these views, which are then ``lifted'' across all other axes automatically.
Concretely, each axis of a Penzai \texttt{NamedArray} has either a position or a name (but not both). Individual named axes can be converted to positional views using \texttt{.untag(...)}, and JAX functions can be vectorized over remaining named axes using \texttt{pz.nx.nmap}; later, the positional view axes can be re-bound to names using \texttt{.tag(...)}. Thus, computations like ``take a softmax over axis \texttt{foo}'' can be expressed as
\vspace{-0.5em}
\begin{verbatim}
  pz.nx.nmap(jax.nn.softmax)(
    array.untag("foo"), axis=0
  ).tag("foo")
\end{verbatim}
\vspace{-0.5em}
without requiring a named-axis-specific implementation of \texttt{softmax}. This means Penzai's named axis system is compatible with the full JAX (and, thus, Numpy) array APIs.

\subsection{Models Are JAX Pytrees, Plus Mutable State}

Similar to Equinox \citep{kidger2021equinox}, each of Penzai's layer classes is immutable and is registered as a JAX pytree\footnote{\texttt{\href{https://jax.readthedocs.io/en/latest/pytrees.html}{jax.readthedocs.io/en/latest/pytrees.html}}}, which makes it possible to manipulate using common JAX utilities. However, mutability can be useful for implementing parameter sharing, per-layer state, and extraction of intermediate activations. To support these use cases, Penzai layers are allowed to store \emph{mutable variables} as attributes, which come in two forms: \texttt{Parameter}s, which are updated by optimizers, and \texttt{StateVariable}s, which are usually updated during each layer's forward pass. Both types of variable can be freely shared between layers.\footnote{Penzai's original ``V1'' neural network system did not store mutable variables in the model, and instead expressed parameter sharing and state using a ``data-effect'' system, which rewrote the model structure when the model was called. This paper describes the newer ``V2'' design, which directly supports mutable state.}

To allow them to be safely used with JAX's function transformations, variables can be ``frozen'' before applying a function transformation, which turns them into ordinary immutable JAX pytrees. These frozen variables can then be temporarily ``unfrozen'' (which creates new mutable copies) inside the transformed function, then frozen again before returning from the function, resulting in a purely-functional view of the model's behavior. Frozen parameters can also be directly embedded into the model once they no longer need to be updated.

\subsection{Selectors Enable Flexible Tree Modifications}
To help users modify Penzai models, Penzai includes a powerful tree-rewriting utility, \texttt{pz.select}, which ``selects'' parts of data structures by type or position. For example, the expression
\vspace{-0.5em}
\begin{verbatim}
  pz.select(model)
    .at_instances_of(pz.nn.Attention)
    .at_instances_of(pz.nn.Softmax)
    .insert_after(some_new_layer)
\end{verbatim}
\vspace{-0.5em}
will return a copy of the model with new logic inserted after each attention pattern computation, making it possible to retrieve or modify these attention patterns. It is similarly possible to e.g. swap out pretrained \texttt{Linear} layers for low-rank finetuning layers using code like
\vspace{1em}
\pagebreak[3]
\begin{minipage}{\linewidth}
\begin{verbatim}
  pz.select(model)
    .at_instances_of(pz.nn.Linear)
    .apply(loraify)
\end{verbatim}
\end{minipage}
\\[1em]
\texttt{pz.select} uses the JAX pytree registry to support modifying any JAX-compatible object.
Because Penzai models are JAX pytrees, \texttt{pz.select} can be used to perform arbitrary modifications to Penzai models. And because Penzai models intentionally expose the model forward pass using combinators, and decompose operations into independent, semantically meaningful chunks, users are free to intervene at arbitrary points and insert logic similar to hook-based approaches. Direct model editing also supports a wider set of transformations than hook points, such as replacing individual model components with linear approximations.

\section{Treescope: Automatic Visualization of Models and Array Data}

Exploratory research with neural networks often involves manipulating deeply-nested data structures containing multidimensional arrays, which can be difficult to summarize and visualize. When using Penzai's model-editing tools to change the structure of a complex model, visualizing the modified structure can also be very useful to confirm that the correct modification was made.

\begin{figure}[t]
    \centering
    \includegraphics[width=0.95\linewidth,trim={0 0 3.5in 0},clip]{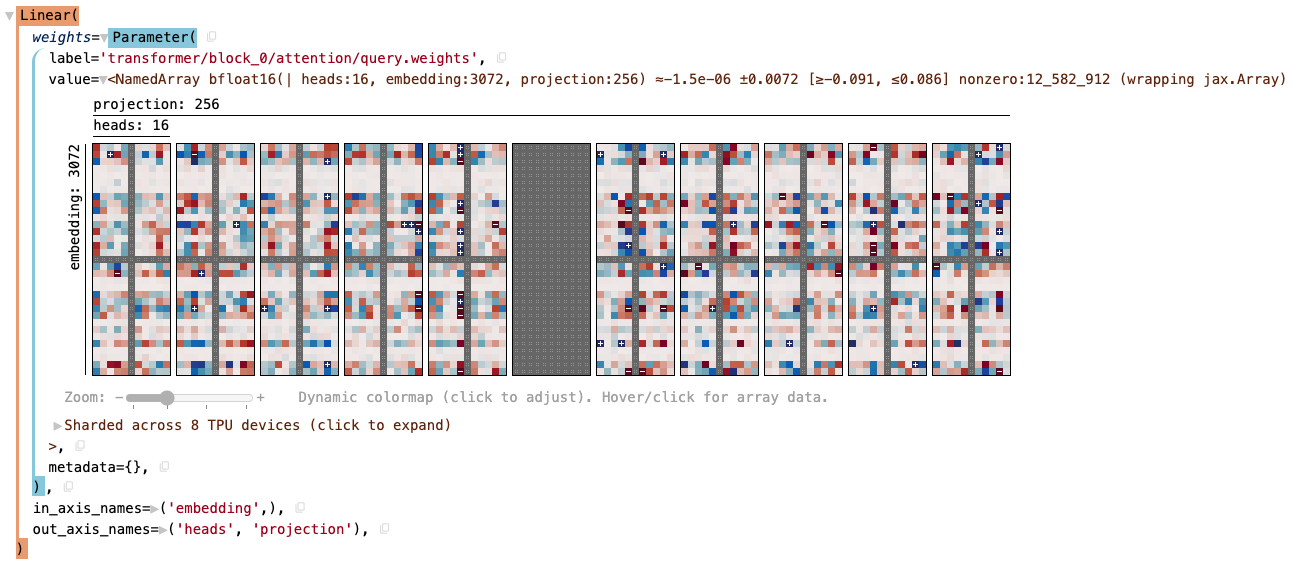}
    \vspace{-0.5em}
    \caption{When pretty-printing a Penzai \texttt{Linear} layer, Treescope renders an inline faceted visualization of the parameter array.}
    \label{fig:treescope-linear}
\end{figure}

The Treescope pretty-printer simplifies this process by automatically producing rich interactive visualizations of deeply-nested machine learning data structures, including both model objects and their inputs and outputs.
Treescope directly integrates with Jupyter IPython notebook environments \citep{kluyver2016jupyter}, and includes an automatic array visualizer, which renders faceted summaries of multidimensional array shapes and values and directly embeds them into the pretty-printed output.
To prevent outputs from becoming too large, Treescope automatically truncates the array contents to show only a small slice (similar to the ordinary \texttt{repr}), but also adds a summary of the distribution of values across the entire array.
An example rendering of a \texttt{Linear} layer with Treescope, including an inline array visualization, is shown in \cref{fig:treescope-linear}.

Users can click Treescope renderings to fold or unfold individual components, allowing them to ``drill down'' into components of interest.\footnote{This interface was inspired by similar interfaces in the JavaScript console in Google Chrome and Firefox.}
Treescope also inserts ``copy path'' buttons at every level of the printed tree, which show how to extract the clicked component from the original object for further manipulation. And in ``roundtrip mode'' (toggled by pressing the \texttt{r} key), Treescope adds fully-qualified names to all classes, making it possible to directly execute the pretty-printed code to rebuild supported data structures.

Treescope and Penzai were designed together, and Penzai's model components are specifically implemented to be easy to visualize and explore in Treescope. In particular, Penzai models support \emph{fully-roundtrippable pretty-printing}: the pretty-printed output of a Penzai model is enough to rebuild the model architecture even after modifications are made by the user. However, Treescope also supports visualizing models and data structures from other neural network libraries. In particular, Treescope supports rendering arbitrary JAX pytrees and models built using the JAX libraries Equinox and \texttt{flax.nnx} (even when model weights are sharded across multiple accelerators), and also supports rendering PyTorch modules and tensors. Treescope also includes an extension system, which makes it possible for users to add rich visualization support for new types, or to replace Treescope's automatic  array visualization with different types of inline figure.
Ultimately, the goal is to allow Treescope's visualization system to be used in combination with other interpretability tools, without having to first rewrite existing models using Penzai.

\section{Using Penzai and Treescope for Interpretability Research}

\subsection{Transformer Implementation}
As a starting point for research into interpreting and controlling model behaviors, Penzai includes a generic Transformer \citep{vaswani2017attention} language model implementation, which uses Penzai's combinators to directly mirror the transformer forward pass in the the model's structure.
This implementation supports loading a variety of pretrained models, including  Gemma 2B and 7B \citep{team2024gemma}, Llama 1, 2, and 3 \citep{touvron2023llama,touvron2023llama2}, Mistral 7B \citep{jiang2023mistral}, and the Pythia scaling suite \citep{biderman2023pythia}. Following Penzai's design conventions, each of these architecture variants corresponds to a different specialization of the same \texttt{TransformerLM} base class (and common components \texttt{TransformerBlock}, \texttt{Attention}, and \texttt{TransformerFeedForward}). Architectural differences are encoded by using different sublayer arrangements, and each model can be freely reconfigured by the user. Capturing or intervening on intermediates, fine-tuning, low-rank adaptation, and sampling are all supported, and can even be combined with each other. Additionally, due to JAX's simple APIs for multi-device computation, all of these modifications can be seamlessly distributed across multiple accelerator devices.

\begin{figure}[t]
    \centering
    \includegraphics[width=0.95\linewidth]{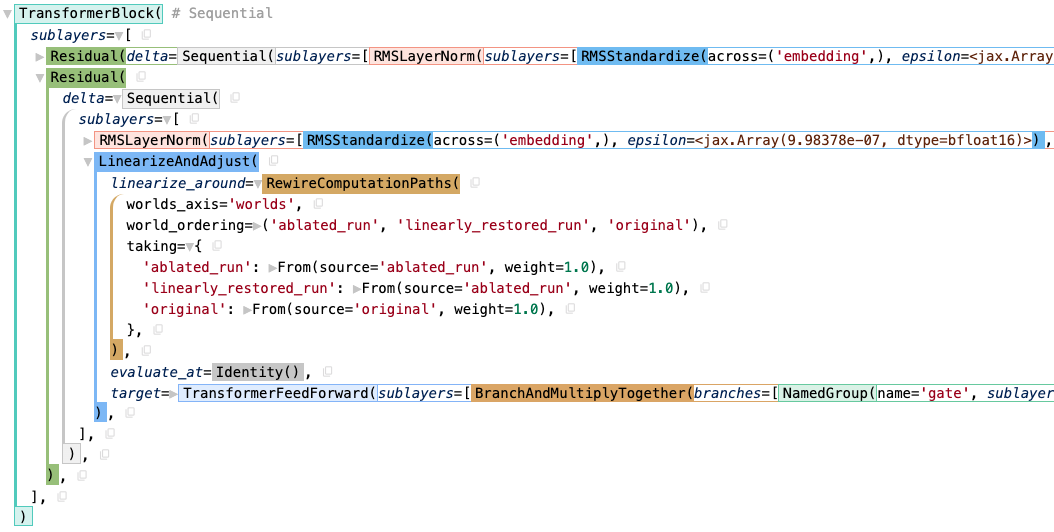}
    \vspace{-0.5em}
    \caption{A modified Transformer block, where the feed-forward layer has been replaced with a \texttt{LinearizeAndAdjust} combinator (which computes a linear approximation of its target layer) and a \texttt{RewireComputationPaths} operation (which copies activations across a named ``worlds'' batch axis).}
    \label{fig:rewiring}
\end{figure}

\subsection{Utilities for Common Operations}
\begin{figure}
    \centering
    \includegraphics[width=0.95\linewidth]{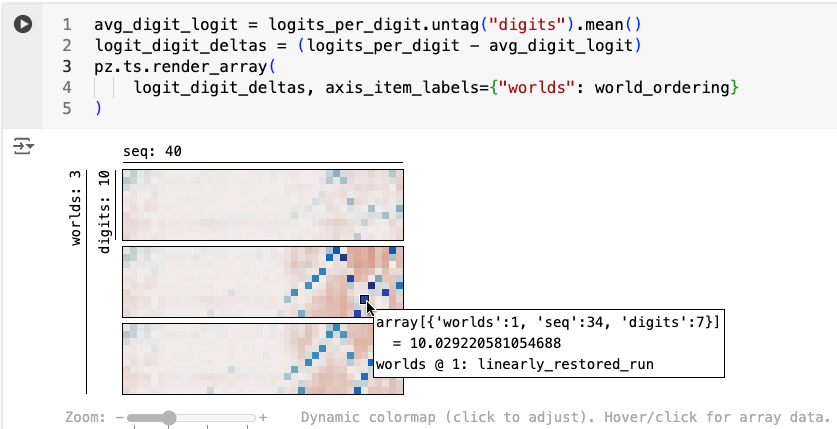}
    \vspace{-0.5em}
    \caption{A visualization of a rank-3 array of logit differences using Treescope (from the intervention in \cref{fig:rewiring}), with a  mouse tooltip giving more information about a specific array element.}
    \label{fig:logit-diffs}
\end{figure}

Penzai includes a collection of extra utilities in \texttt{penzai.toolshed}, including tools for patching individual activations between counterfactual model inputs \citep[see][]{zhang2023towards}. Unlike hook-based workflows, these tools work by inserting new layers into the model, resulting in a copy of the model that includes the given intervention, and avoiding the need to manage global state. Directly editing the model structure also enables interventions that cannot be expressed easily with hooks, such as linearizing parts of a model for easier analysis (shown in \cref{fig:rewiring,fig:logit-diffs}).

\texttt{penzai.toolshed} also includes utilities for basic training, low-rank finetuning, shape annotation, and multi-device sharding. Each utility has a self-contained implementation as a model transformation, and can either be used as-is or taken as a starting point for more complex workflows.

\subsection{Example: Finding Induction Heads In Gemma 7B}

Penzai includes a tutorial notebook walking users through the process of analyzing the Gemma 7B open-weights model \citep{team2024gemma} in a Colab notebook%
\footnote{\url{https://penzai.readthedocs.io/en/stable/notebooks/induction_heads.html}},
starting with exploring the model's structure and predictions on simple examples, then visualizing attention patterns throughout the model to identify candidate induction heads, and finally ablating and patching them to confirm that they are responsible for the copying behavior, while using Treescope to get quick interactive feedback throughout the process.

Because it uses JAX as a backend, exploration in Penzai can immediately benefit from many of JAX's features. Each step in the notebook is seamlessly parallelized across multiple TPU devices. Additionally, the effects of individual MLP layers can be decomposed into linear and nonlinear components using JAX's linearization transform \texttt{jax.jvp}.

\section{Discussion}
By representing models as compositions of simple building blocks, and providing powerful tools for both visualizing and editing these data structures, Penzai and Treescope can support a wide variety of use cases without restricting expressivity. These libraries are under continued development, and I hope they can serve as useful tools for new research on interpreting machine learning models, understanding their training dynamics, and steering their behaviors.

\section*{Acknowledgements}
Building Penzai and Treescope would not have been possible without the help of Dougal Maclaurin, who supported the open-source release and gave useful feedback throughout the development process. I would also like to thank Danny Tarlow, Hugo Larochelle, David Duvenaud, and Chris Maddison for their advice and encouragement.

\bibliographystyle{icml2024}
\bibliography{main}

\onecolumn
\newpage
\appendix

\section{Additional Penzai and Treescope Visualizations}

\begin{figure}[h!]
    \centering
\includegraphics[width=\linewidth,trim={0 2.75in 17in 0},clip]{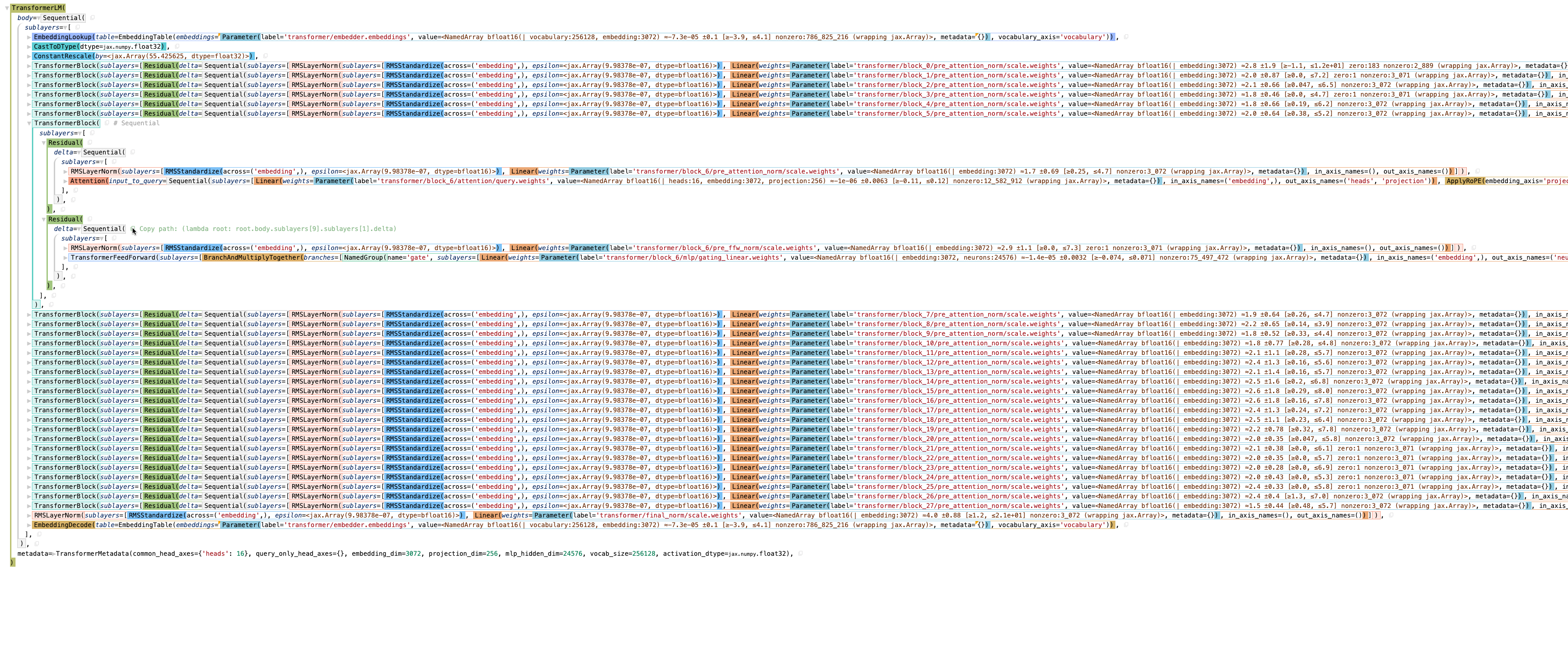}
    \caption{The Gemma 7B open-weights model \citep{team2024gemma}, loaded using Penzai's transformer implementation and visualized using Treescope. The mouse cursor is hovering over a ``copy path'' button, which copies the location of the selected object to the clipboard when clicked.}
    \vspace{-2em}
    \label{fig:app-gemma}
\end{figure}

\begin{figure}[b!]
    \centering
    \includegraphics[width=0.475\linewidth,trim={0 0 6in 0},clip,valign=t]{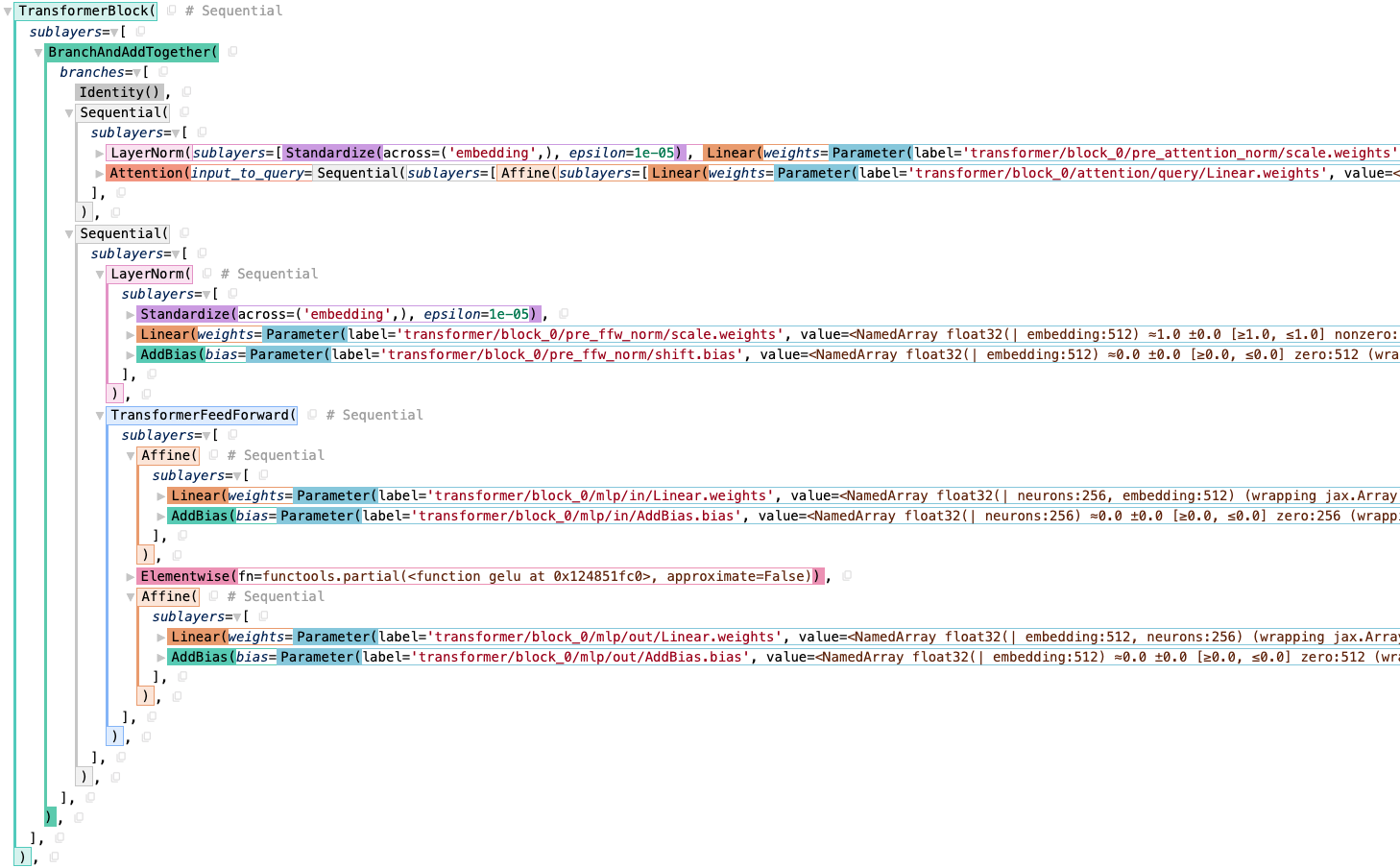}
    \hspace{0.25cm}
    \includegraphics[width=0.475\linewidth,trim={0 0 6in 0},clip,valign=t]{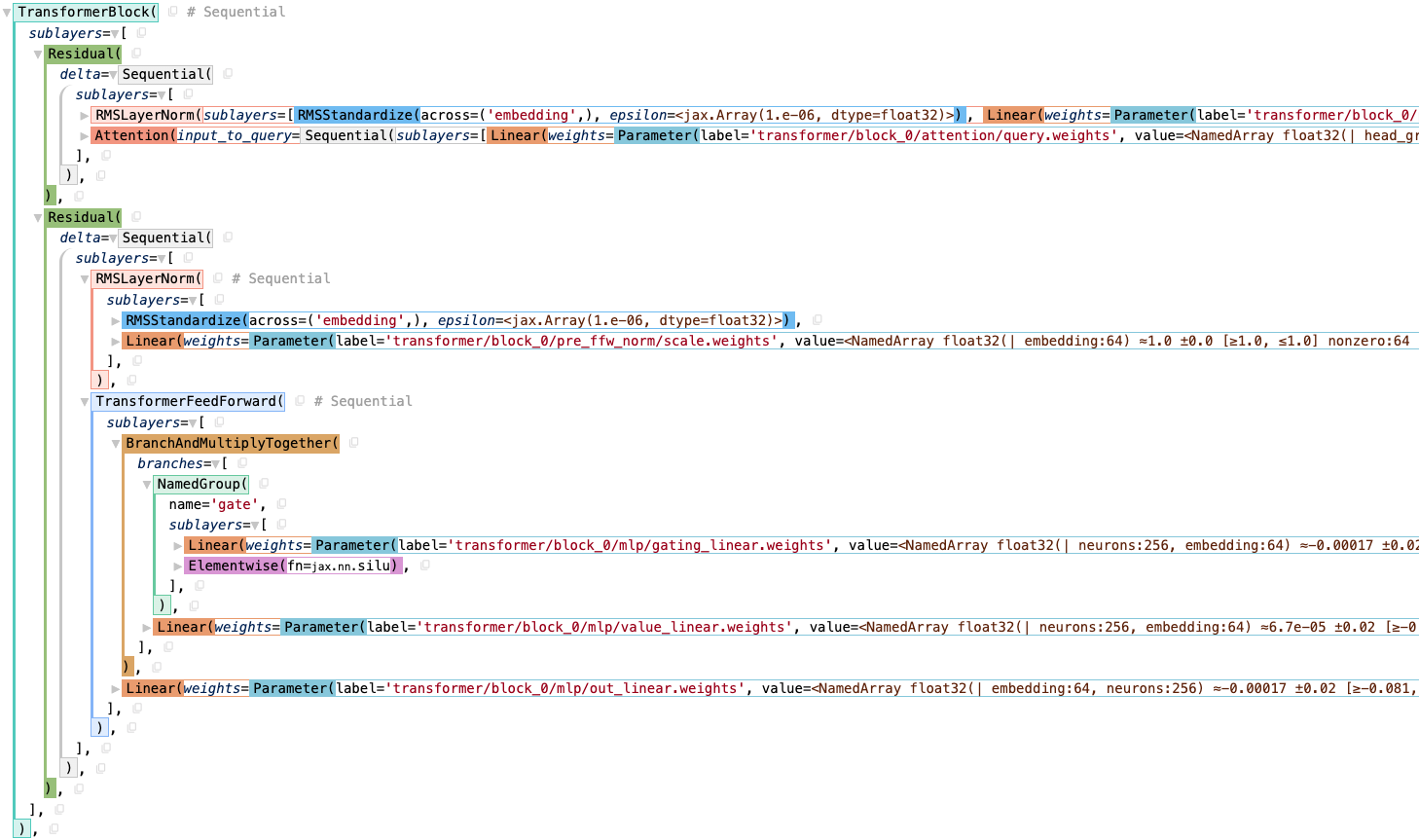}
    \caption{Comparison of Transformer block structures for GPT-NeoX/Pythia (\citealt{biderman2023pythia}, left) and Llama (\citealt{touvron2023llama}, right) architectures. GPT-NeoX runs the attention and feedforward parts in parallel, and uses LayerNorm \citep{ba2016layer} and a standard MLP network with biases. In contrast, Llama uses two separate residual blocks, and uses RMSNorm \citep{zhang2019root} and a gated feedforward network \citep{shazeer2020glu} with no learned bias terms. These differences can be concisely expressed using different arrangements of Penzai combinators. (Any class with a ``\texttt{\# Sequential}'' annotation simply runs its children in order, without custom logic. Subclasses of \texttt{Sequential} are often used to improve readability and allow manipulation by \texttt{pz.select}.)}
    \label{fig:app-variants}
\end{figure}

\begin{figure}[p]
    \centering
    \includegraphics[width=\linewidth]{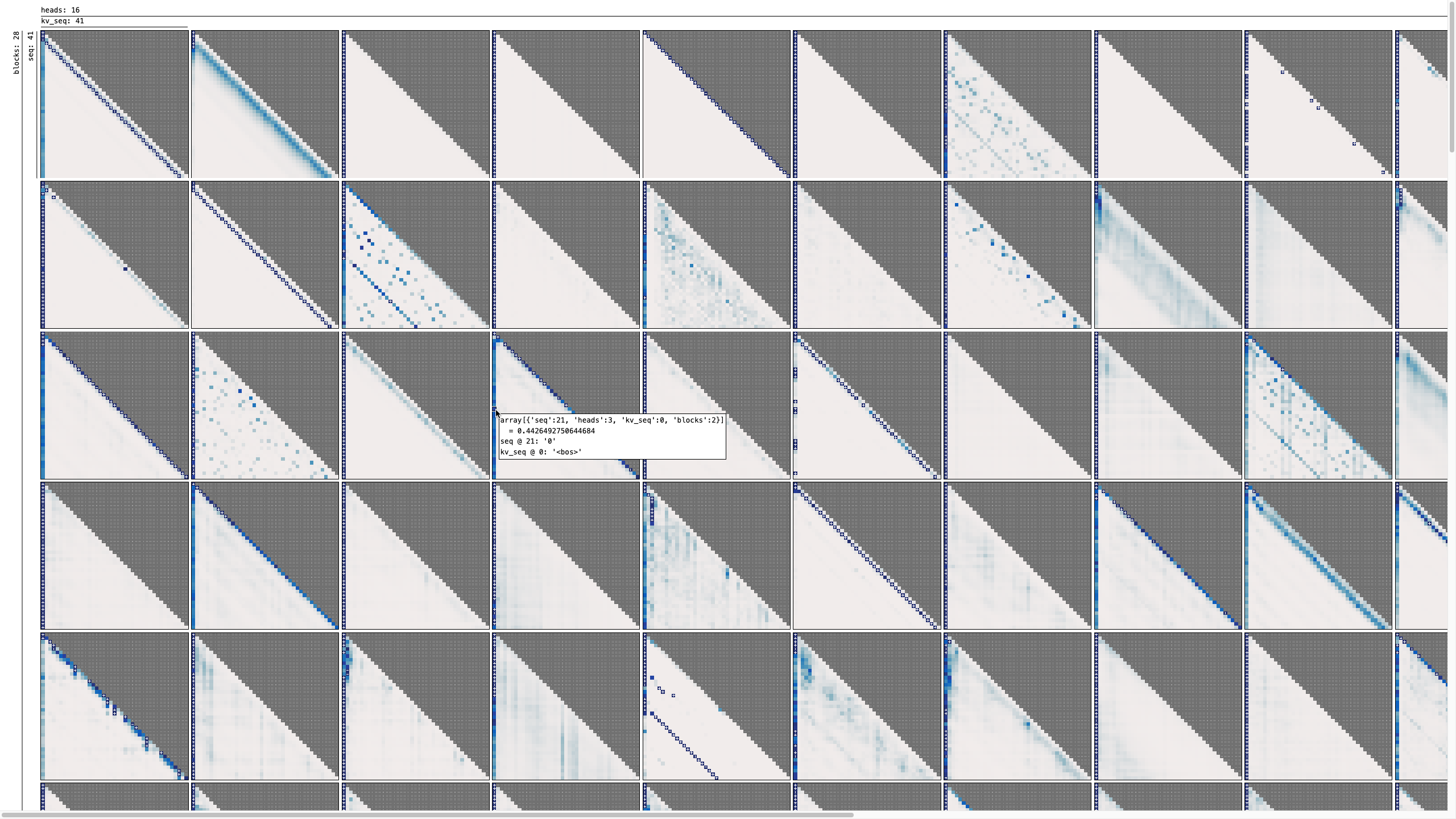}
    \vspace{-1.5em}
    \caption{A faceted rendering of attention patterns on an example sequence, produced by Treescope. The hover tooltip gives information about the specific tokens being attended to by the head under the mouse cursor.}
    \label{fig:app-attn}
\end{figure}

\begin{figure}[p]
    \centering
    \includegraphics[width=0.8\linewidth]{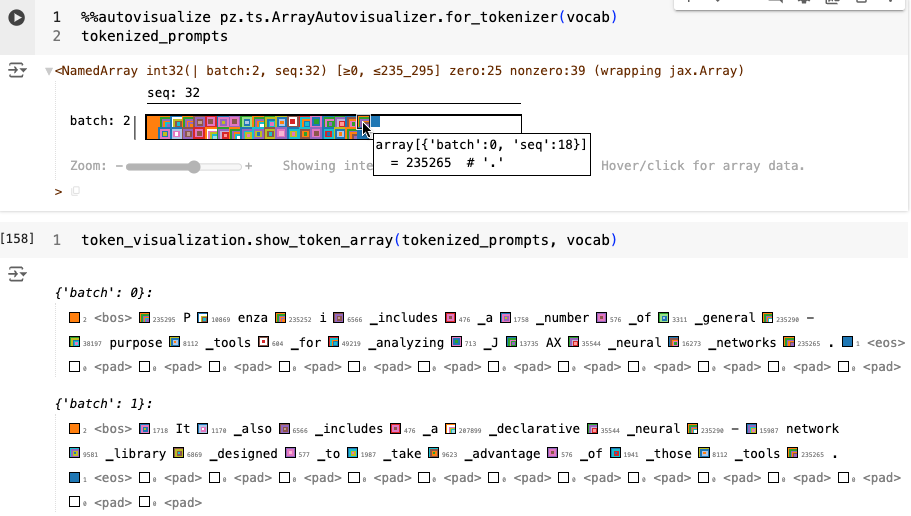}
    \caption{Two renderings of a batch of token sequences. The first uses Treescope's default array visualizer for discrete data, which maps each token ID to a unique ``digitbox'' pattern, where each color corresponds to one digit of the value. The second interleaves these with the token values, using a token-visualization helper function. Control tokens and padding are easily recognizable across the two sequences due to having single-digit token IDs, which map to solid-color box renderings.}
    \label{fig:app-tokens}
\end{figure}

\end{document}